# Incremental Classifier Learning Based on PEDCC-Loss and Cosine Distance

Qiuyu ZHU,　Zikuang HE,　Xin YE
School of Communication & Information Engineering, Shanghai University

## 1 Abstract：

The main purpose of incremental learning is to learn new knowledge while not forgetting the knowledge which have been learned before. At present, the main challenge in this area is the catastrophe forgetting, namely the network will lose their performance in the old tasks after training for new tasks. In this paper, we introduce an ensemble method of incremental classifier to alleviate this problem, which is based on the cosine distance between the output feature and the pre-defined center, and can let each task to be preserved in different networks. During training, we make use of PEDCC-Loss to train the CNN network. In the stage of testing, the prediction is determined by the cosine distance between the network latent features and pre-defined center. The experimental results on EMINST and CIFAR100 show that our method outperforms the recent LwF method, which use the knowledge distillation, and iCaRL method, which keep some old samples while training for new task. The method can achieve the goal of not forgetting old knowledge while training new classes, and solve the problem of catastrophic forgetting better.

## 2 Introduction：

As is known to us all, human vision system is essentially incremental, new visual information can be obtained under the premise of preserving the old visual information. For Instance, man may be able to recognize more kinds of vehicles after visiting a car exhibition while still can recognize the car in his garage. Nowadays most pattern recognition system's training data is divided into mini-batches during training. The dataset's label is available and they are trained in a random order. However, with the rapid development of computer vision and artificial intelligence, we need a more flexible strategy to deal with the large amount of information with dynamic properties in real life. At least we should achieve class increment in the field of image classification, namely the class-incremental learning. So, an incremental method is required under this goal to keep learning new classes without forgetting what the network has learned.

The point of studying incremental learning is twofold: on the one hand, with the rapid development of Internet information technology, massive data are accumulated in various fields. And we may not preserve the old model's training data due to the copyright and limited storage space. So how to effectively and continuously obtain the valuable information from these datasets is an important research direction, incremental learning can solve the demand which traditional training strategy cannot. On the other hand, researchers can have a better view of the biological neural networks and the way human brain learns by studying incremental learning, which provides a solid theoretical and technical basis for the creation of new computational models and efficient learning algorithms.

However, when a convolutional neural network is trained on different classes of data in different time, the network will lose the ability to get good features of old data in previous tasks, which is called catastrophic forgetting[1]. So, the biggest challenge for incremental learning right now is the catastrophic forgetting, which means that we should balance knowledge of new classes with knowledge of old classes. Quote the definition given by *Polikar*[2], an incremental learning algorithm should maintain the knowledge learned in the past and cannot re-process the processed data.

Traditional methods of incremental learning like ensemble learning[3] trains multiple models first, and then combine them together to achieve a unified model to have an accurate, stable and robust model. Current incremental learning studies make more use of convolutional neural networks, such as LwF[4], which propose a method for training incremental convolutional neural network, in which each class shares the same convolution layer, and the only difference is the classifier. When some new classes come, a new classifier is added for the new classes, and distillation loss function is used for network fine-tuning to preserve the old classes' knowledge. Hear the loss function is the key to preserve the old tasks' knowledge. iCaRL[5] proposes to retain some examples of previous task to help with training for new tasks, and use the NME classification and distillation loss function as well.

In this paper, we will introduce a method based on cosine distance and ensemble learning to achieve class-incremental learning in the strict sense. We use PEDCC-Loss[6] and the linear layer fixed by pre-defined centers[7] of each class to let the features of different classes converge on their pre-defined center. The new network and old network do not train together. By adding new networks for new tasks, we achieve the class-incremental learning.

Frist of all, each network train for their own classes data individually with PEDCC-Loss, they do not interfere each other and their training data are completely separated. Each network has well preserved the feature extraction ability of corresponding training data and the PEDCC-Loss let the feature of different classes in the same network be separated as far as possible, while having a small cosine distance from corresponding center. Different from some current methods of utilizing the old tasks' data, we do not use the data of the previous classes and previous networks at all every time when we learn the new task, thus realizing the real class-incremental learning. The system diagram is shown in figure1 and figure2, corresponding to the training phase and testing phase respectively.

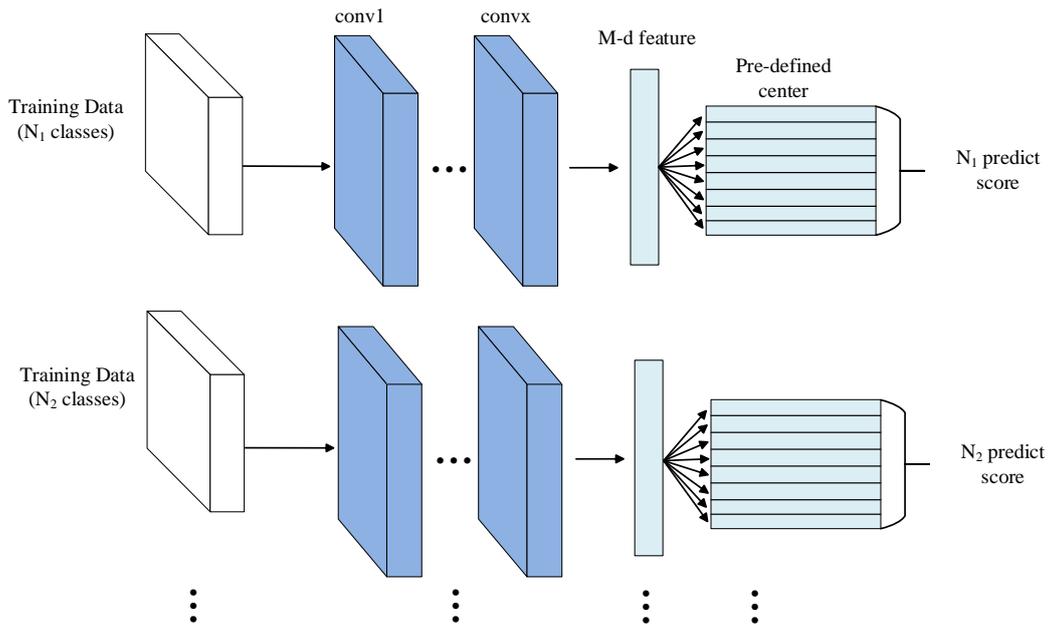

Figure1: Class-incremental learning, an algorithm can achieve class-incremental learning without forgetting the old tasks' knowledge by adding new networks training for new tasks.

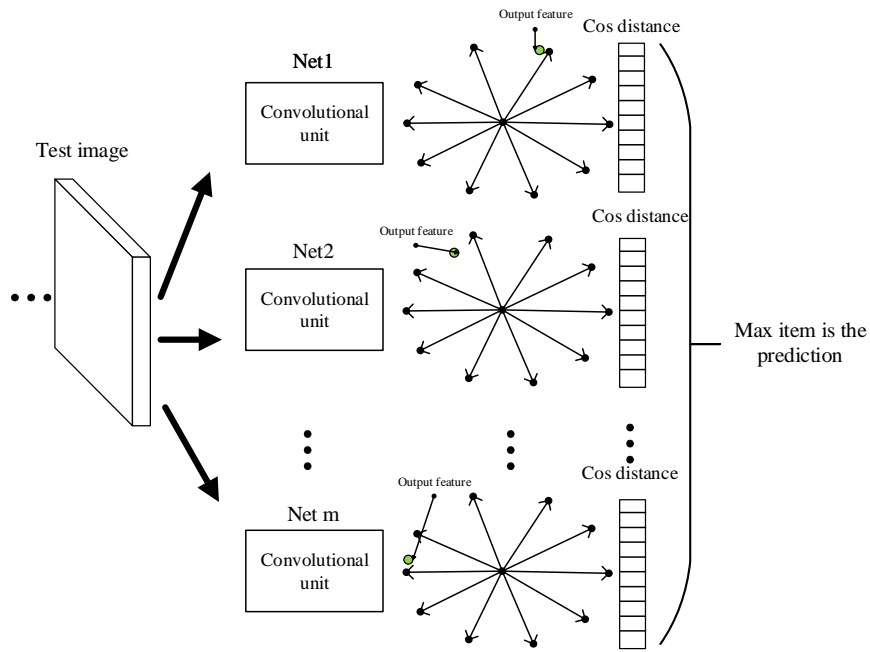

Figure2: testing phase. The test image gets the output feature through each pre-trained network, the cosine distance is calculated between the output features of each network and the pre-defined center used in the training process of the network. The label of pre-defined center which has the largest cosine distance to the output features is the final prediction.

# 3 Related work:

Xiao *et al.*[7] propose a kind of network that can grow in a hierarchical way, each node is composed of some clusters of similar classes. By the tree structure, when the model updates we only need to adjust some parts of the model, and the model adjustment scope can be strictly controlled. Incremental learning is realized through the growth of network, but the program is faced with the increasing difficulty of training large network and the difficulty of how to effectively increase network capacity. For the catastrophic forgetting problem in incremental learning of convolutional neural networks, Rusu *et al.*[8] present progressive NN, used to solve the problem of networks adapting to new classes (increments).The idea is to keep all the networks of the previous class, create a new network for each new class, while retaining the low-level characteristics of the old networks. Venkatesan *et al.*[9] introduce Phantom Sampling produced by GAN to retain the information of the original training samples. These Phantom samples are used to train the new deep network together with incremental samples, achieving a good result of incremental training of classes. However, this method takes a long time to train and is difficult to be applicable to the new incremental samples of old classes.

Zhizhong Li *et al.*[4] propose a method called Learning without Forgetting, using the distillation loss function, classification loss function and fine-tuning to retain the original model knowledge in training classes of new tasks, the role of the distillation loss function is to make a new class of the output of the network model in the original class is more approaching to the original network so as to keep the original network learned information, by setting the proportion of the classification loss to the distillation loss you can control whether the data is more biased towards the original or the new class. Rebuffi et al.[5] suggest that the convolutional neural network is used for feature learning and representation. The new classes samples and the previously stored old class sample are added to the convolutional neural network for training together, so as to update the current model parameters and obtain new feature representation for all classes. In classification, Using the idea of NCM [11] for reference, Nearest-Mean-of-Exemplars is used to classify the feature vectors extracted from the sample set after averaging them. Wu[12] et al. redefined the loss function (cross entropy loss function + distillation loss function) on the basis of iCaRL, and added GANS[13] to generate a few samples of the old class, which improving the generalization ability. Xin Ye et al.[14] present a method to do class-incremental learning based on one class classifier SVDD, where AM-softmax is used to train convolutional neural network as the feature extractor network of old task's data, then the old task's feature and new task's feature are used to classify incrementally according to SVDD.

# 4 Cosine distance classification based on PEDCC-Loss：

## 4.1 PEDCC-Loss

PEDCC-Loss[15] is a classification Loss function proposed by us, which is based on the pre-defined center, and can make the features of different classes have the largest inter-class distance and the smallest intra-class distance, so as to achieve the SOTA classification performance. Different from traditional loss function used in the process of training neural network, the center of each class's feature is uncertain, while PEDCC-Loss has pre-defined centers for the features of each class, and the pre-defined centers are even distributed on feature hypersphere. Through the weight-fixed last linear layer and improved Softmax loss function, we can make the features of each classes distribute evenly on a hypersphere. The equation is as follows:

$$L_{PEDCC-AM} = -\frac{1}{N}\sum_i log \frac{e^{s \cdot (cos\theta_{y_i} - m)}}{e^{s \cdot (cos\theta_{y_i} - m)} + \sum_{j=1, j \neq y_i}^{c} e^{s \cdot cos\theta_j}} \quad (1)$$

$$L_{PEDCC-MSE} = \frac{1}{2}\sum_{i=1}^{m}\|x_i - pedcc_{y_i}\|^2 \quad (2)$$

$$L = L_{PEDCC-AM} + \sqrt[n]{L_{PEDCC-MSE}} \quad (3)$$

Where Eq. (1) is the AM-Softmax Loss[14], Eq. (2) is the MSE Loss function between output feature and pre-defined center, PEDCC-Loss is obtained by combining them two, n>=1 is a superparameter that can be adjusted.

By using this loss function, we can artificially control the feature distribution of each class. The traditional Softmax loss function only adjusts the classification layer parameters according to the current training data, and does not normalize the weight and features of the classification layer. It is inevitable to be affected by the weight and feature modulus when calculating the cosine distance between features and pre-defined centers, thus reducing the predicting accuracy. Using PEDCC-Loss we can make the features in the feature space have large distance between the classes, and small inner-classes distance. Through the fixed classification layer, the network's final prediction only depends on the output feature and the pre-defined center's cosine distance.

Figure 2 shows the features distribution learned by the convolutional neural network with different Loss functions, where the PEDCC-Loss distribution is the feature distribution after the convolution layer output, and before the normalization and linear classification layer. It can be seen that PEDCC-Loss let the features of each class converge closely, and at the same time there are the largest distances between each class after feature normalization. After feature normalization, the modulus of cluster centers of each class is the same and dispersed sufficiently.

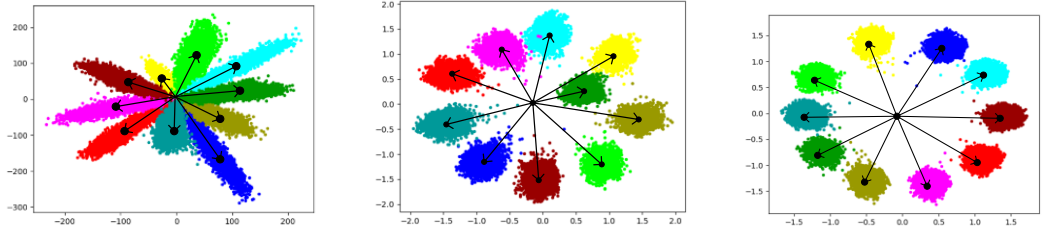

Figure2: the feature distribution of each class through a neural network trained by different loss function. For simplicity, we set the output feature to 2-dimension.

## 4.2 Classification method based on cosine distance

Through the use of PEDCC-Loss, the linear classification layer of the convolutional neural network we trained is fixed. Moreover, the pre-defined center of PEDCC is also the class prototype for the final judgment. Which means for classification, correlation is calculated between the corresponding PEDCC center and the network latent feature, namely the classification result is depend on the correlation (that is cosine distance) between the latent feature and the pre-defined centers. In the linear classification layer, $\mathbf{w}_i$ is the weight of a specific class $i$ defined by PEDCC's pre-defined center, and the output feature $\mathbf{x}$ and $\mathbf{w}_i$ are normalized. According to the formula of the fully connect layer, we have:

$$g_i(\mathbf{x}) = \mathbf{w}_i \cdot \mathbf{x} = ||\mathbf{w}_i||\,||\mathbf{x}||\cos\theta = \cos\theta \qquad (4)$$

where $g_i(\mathbf{x})$ is the correlation of $\mathbf{x}$ and $\mathbf{w}_i$, and is equivalent to the posterior probability that the latent feature $\mathbf{x}$ belongs to class i, that is, it is a discriminant function. The prediction results will depend entirely on the cosine distance between the output latent feature and the predefined center. Then, the classification results can be obtained by maximizing distance:

$$j = \underset{i}{\mathrm{argmax}}\, g_i(\boldsymbol{x}) \qquad (5)$$

This classification result is consistent with the direct network classification output, but because we turn the classification problem into the problem of finding the extremum of the discriminant function, it makes it easy for us to use in the incremental classifier. This classification method is not only simple in calculation, but also more accurate than the SVD-based method [14]. The specific calculation method is shown in algorithm 1:

**Algorithm 1** Incremental classifiers

**Input:** pre-trained network Net1 of M classes

Image set $I=\{I_0,…, I_n\}$ of m classes (M≥m) //m is a subset of M

**Require:** $W=(W_0,…, W_m)$ // the pre-defined centers of m classes

  **for** each image $I_i$ in $I$ **do**

    $x=Net1(I_i)$     //x is the Net1's latent feature of image $i$

    **for** $j$=0→m **do**

      $g_i(x) = W_j \cdot x$

    **end for**

    Output = $\mathrm{argmax}_i(g_i(x))$   // result depends on cosine distance

  **end for**

## 5 Incremental classification of single trained multi-class network

With the above classification method based on cosine distance, for a trained multi-class classification network, we can classify any two or more classes' samples. This can be used in limited scenarios where only a limited number of class samples need to be classified. For example, for the classifier of character, if we only need to recognize the number 0-9, then we only need to discriminate in 10 cosine distances.

Taking the Cifar100 dataset as an example, a classification network is firstly trained with 100 data classes, then ten of the 100 classes were randomly selected, and the features were obtained through the trained network, multiplied by the weight of the corresponding classification layer of the ten classes. At this time, the accuracy of classification of 10 can reach 91.8%, which is much higher than the accuracy of classification of 10 data through the original network, which is 78.05%. By the same analogy, the data of 20 and 30 classes' accuracy can be improved in the way of incremental classifier without changing the original network. A specific result is shown in figure 3.

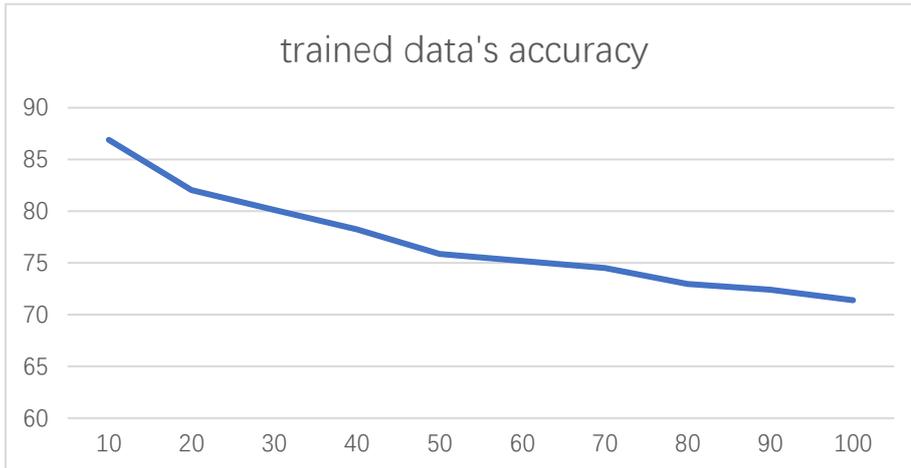

Figure 3: A result of incremental classification of single trained multi-class network

## 6 Incremental learning with multiple classification networks

The main purpose of previous incremental learning methods like LwF, iCaRL, is to preserve the knowledge of old task as much as possible, and at the same time to learn new knowledge of new task. Different from the above methods, when some new classes' data arrive, we use them to train a new network with PEDCC-Loss. In this way, different training data's knowledge is saved in different networks. Finally, the combination of classifier is realized by cosine distance.

Firstly, set the number of classes for increment according to the specific scenario, let's call the number N$i$. Then produce pre-defined center for each class by PEDCC algorithm and set the pre-defined centers' dimension M at the same time. Finally, a PEDCC matrix of N rows M columns is produced to be the classification layer weight of the neural network. For instance, when $N_1$=10, M=512, the number of initial data classes is 10, which means the training data of the first network is those data of 10 classes. The first neural network uses this matrix with $N_1$ rows and M columns as the linear layer weight, and retain the network model after training. The PEDCC centers are more than the weight of classification layer, they also serve as the labels of each class.

When new classes appear, we create PEDCC centers to train second neural network, if the number of new data's class is the same as one network trained before, we can use the same PEDCC center, otherwise we produce new centers according to the new class number. In this step, the old training data is not used at all. Neural network is trained on the cosine distance between the output feature and the pre-defined center. After training, the second network is also saved, and we do the same thing when other classes of data arrived.

The test process is shown in algorithm 2. We trained all classes by different networks, making the distribution of the features of each class are close to their pre-defined center. The test set samples pass through trained networks to get the feature of each network, then calculate the cosine distance between the output feature and the pre-defined centers, to select the pre-defined center whose cosine distance is the largest from the output feature and use the label of the center as the final prediction.

**Algorithm 2** Testing Process

**Input:** test image set $I=\{I_0,…, I_n\}$ of all trained classes

**Require:** $N=(N_0,…, N_m)$ // total pre-trained networks

$W=(W_0,…, W_m)$ // the pre-defined center of corresponding network

for each image $I_i$ in $I$ do

    for each net $N_j$ in $N$ do

        $x_j = N_j(I_i)$  // x is the latent feature

        $g_j(x) = W_j \cdot x_j$

        $S_j = max_j(g_j(x))$  // $S_j$ is the prediction score of $N_j$

    end for

    $S = \{S_0, …, S_j\}$

    $o = argmax_j(S)$  // choose the network

    $output = N_o(I_i)$  // final prediction

end for

# 7 Experimental results

The experimental data set used in this paper are EMNIST[15] with 40 classes and CIFAR100[16] with 100 classes. For EMNIST, we choose the modified VGGNet[17] as our classification network for EMNIST, multiple convolutional neural networks with the same structure train for their own training sets. For CIFAR, we choose the WideResnet[18] as our network structure. The network structure is shown in table1 and table2:

Table1: Network Structure for Incremental Classification of EMNIST

| Layer | EMNIST |
| --- | --- |
| Conv0.x | [3×3, 64] ×1 |
| Conv1.x | [3×3, 64] ×3 |
| Pool1 | 2×2 Max, Stride 2 |
| Conv2.x | [3×3, 128] ×3 |
| Pool2 | 2×2 Max, Stride 2 |
| Conv3.x | [3×3, 256] ×3 |
| Pool3 | 2×2 Max, Stride 2 |
| Fully Connected | 512 |
| Fixed Fully Connected | 40 |

Table2: Network Structure for Incremental Classification of CIFAR100

| Group name | Block type=$B(3,3)$ |
| --- | --- |
| Conv1 | $[3 \times 3, 16]$ |
| Conv2 | $\begin{bmatrix} 3 \times 3, 16 \times 10 \\ 3 \times 3, 16 \times 10 \end{bmatrix} \times 4$ |
| Conv3 | $\begin{bmatrix} 3 \times 3, 32 \times 10 \\ 3 \times 3, 32 \times 10 \end{bmatrix} \times 4$ |
| Conv4 | $\begin{bmatrix} 3 \times 3, 64 \times 10 \\ 3 \times 3, 64 \times 10 \end{bmatrix} \times 4$ |
| Avg-pool | $[7 \times 7]/[8 \times 8]$ |
| Fully Connected ip1 | 100 |
| Fully Connected ip2 | Softmax |

The size of the convolution kernel is uniform in convolution layers, which is 3×3, and the stride and padding is 1. We use the pytorch1.0 framework to train our neural networks for 120 epochs. The learning rate starts at 0.1 and is divided by 10 after 20,50,80,100 epochs. We train the networks using SGD with batch size 256, a weight decay parameter of 0.0005 and the momentum is 0.9. Because of the MNIST and CIFAR10 dataset's classes are 10, which are too small to reflect the effect of class-incremental learning, so we use the EMNIST and CIFAR100 to do our experiment.

In the test phase, one image pass through each pre-trained network and we have the features of all networks. Then we calculate the cosine distance between the output feature and the corresponding pre-defined centers, the pre-defined center which has the maximum distance is the final prediction. The testing process is shown in figure 2.

## 7.1 EMNIST experimental result

This section shows the performance of our method in EMNIST. We use two or three networks to do the incremental learning, the capacity of each network varies from 10 to 30 classes. The classification accuracy of each batch and the total classes' accuracy is also given.

Table 3: The result of class-incremental experiment on EMNIST

| Test classes | One batch | Two batches | Three batches |
|---|---|---|---|
| | Accuracy | Accuracy | Accuracy |
| 10+10 (2networks) | 99.45% | 97.71% | / |
| 20+20 (2networks) | 99.14% | **97.5%** | / |
| 30+10 (2networks) | 98.89% | 96.75% | / |
| 20+10+10 (3networks) | 99.11% | 95.18% | 94.05% |
| CNN+Softmax | **98.24%** | | |

The accuracy of 40 classifications based on cross-entropy loss function is 98.24%. From the table, we can see that the 40 classifications accuracy of 20 + 20 of the two networks can reach 97.5%, which is very close to the result of direct training with softmax loss function. When we use three networks instead of two networks to train incrementally, the final performance declined. Because with the increase of the number of networks, more interference factors will inevitably be brought in the feature selection stage. However, it can be seen that under different combinations, the overall recognition rate of 40 classes is maintained at a higher level.

## 7.2 CIFAR100 experiment result

This section uses the same strategy as is mentioned before, we divide the CIFAR100 into different combinations, and increased to five networks, the result is shown in table 4.

Table 4: The result of class-incremental experiment on CIFAR100

| Test classes | One batch | Two batches | Three batches | Four batches | Five batches |
|---|---|---|---|---|---|
| | Accuracy | Accuracy | Accuracy | Accuracy | Accuracy |
| 50+50 (2networks) | 81.57% | **69.26%** | / | / | / |
| 80+20 (2networks) | 79.25% | 69.8% | / | / | / |
| 40+30+30 (3networks) | 83.18% | 69.99% | 63.04% | / | / |
| 25+25+25+25 (4networks) | 86.54% | 71.56% | 64.19% | 59.17% | / |
| 20+20+20+20+20 (5networks) | 86.91% | 74.98% | 67.25% | 61.41% | 57.07% |
| CNN+Softmax | **76.25%** | | | | |

The accuracy of 100 classifications based on cross-entropy loss function is 76.25%. From the table, we can see that the 100 classifications accuracy of 50+50 of the two networks can reach 69.26%. As the number of networks increases, accuracy decreases, which is the same as the conclusion in the previous section. The incremental classification of 100 is also close to the traditional classification result. It shows that the integrated incremental network can achieve real incremental learning classification with a fair good result.

The experimental results in this paper were compared with some other methods on cifar-100, as shown in the figure 6 below. As we can see from it, our method is better in absolute value of recognition accuracy. The recognition accuracy decreases with the increase of the number of classes, which is consistent with the iCaRL method which use some old samples, and is obviously better than the LwF method using knowledge distillation.

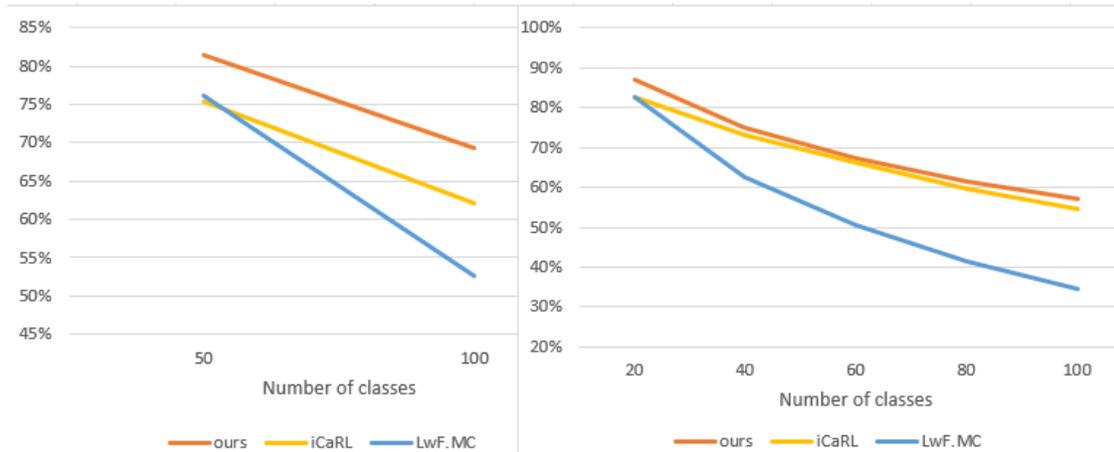

Figure 6: Experimental results of class-incremental training on CIFAR100. The left picture is the 50+50 incremental result, and the right is the 20+20+20+20+20+20 incremental result.

# 8 Conclusion and discussion

The field of incremental learning is still full of unknowns and challenging. Some current incremental learning algorithms use network distillation, retaining a small number of old samples and generating old samples through GAN to learn new knowledge and try not to forget the knowledge previously learned.

This paper proposes a incremental learning method based on Independent multiple network learning, in the training progress for new classes the old task's sample are not used totally, and the PEDCC-Loss is combined to make different class's feature have a small cosine distance from the pre-deifned center. The experiments of EMNIST and CIFAR100 datasets show a good results.

However, there are still some areas to be improved in the algorithm. For example, when the number of networks increases, the final performance decreases due to the introduction of more interference. We expect to use knowledge distillation, multi-network fusion training and other methods to let test image belong to their corresponding class as much as possible. In this way, the final recognition accuracy will have some room for improvement.